\documentclass[letterpaper, 10 pt, conference]{ieeeconf}  % Comment this line out if you need a4paper

\IEEEoverridecommandlockouts                              % This command is only needed if 
\usepackage[dvips]{graphicx}
\usepackage{amsmath, amssymb}
\usepackage{bm}
\usepackage{comment}
\usepackage{color}
\usepackage{multirow}
\usepackage{cite}

\title{\LARGE \bf
Reinforcement Learning of Multi-robot Task Allocation for Multi-object Transportation with Infeasible Tasks
}

\author{Yuma Shida$^{1}$, Tomohiko Jimbo$^{1}$, Tadashi Odashima$^{1}$, and Takamitsu Matsubara$^{2}$% <-this % stops a space
\thanks{$^{1}$The authors are with R-Frontier Division, Frontier Research Center, Toyota Motor Corporation, 1, Toyota-cho, Toyota, Aichi, Japan
        {\tt\scriptsize \{yuma\_shida,         tomohiko\_jimbo, tadashi\_odashima\}@mail.toyota.co.jp}}%
\thanks{$^{2}$Takamitsu Matsubara is with the Graduate School of Science and Technology, Division of Information Science, Nara Institute of Science and Technology, Nara, Japan
{\tt\small  takam-m@is.naist.jp}}%
}

\begin{document}

\IEEEoverridecommandlockouts
{\begin{minipage}{\textwidth}\ \\[18pt] 
  \copyright~2025 IEEE. Personal use of this material is permitted. Permission from IEEE must be obtained for all other uses, in any current or future media, including reprinting/republishing this material for advertising or promotional purposes, creating new collective works, for resale or redistribution to servers or lists, or reuse of any copyrighted component of this work in other works. 
  
  Citation information: DOI 10.1109/SII59315.2025.10870902, 2025 IEEE/SICE International Symposium on System Integration (SII)
\end{minipage}}

\maketitle

\thispagestyle{empty}
\pagestyle{empty}

%%%%%%%%%%%%%%%%%%%%%%%%%%%%%%%%%%%%%%%%%%%%%%%%%%%%%%%%%%%%%%%%%%%%%%%%%%%%%%%%
\begin{abstract}
Multi-object transport using multi-robot systems has the potential for diverse practical applications such as delivery services owing to its efficient individual and scalable cooperative transport. However, allocating transportation tasks of objects with unknown weights remains challenging. Moreover, the presence of infeasible tasks (untransportable objects) can lead to robot stoppage (deadlock). This paper proposes a framework for dynamic task allocation that involves storing task experiences for each task in a scalable manner with respect to the number of robots. First, these experiences are broadcasted from the cloud server to the entire robot system. Subsequently, each robot learns the exclusion levels for each task based on those task experiences, enabling it to exclude infeasible tasks and reset its task priorities. Finally, individual transportation, cooperative transportation, and the temporary exclusion of tasks considered infeasible are achieved. The scalability and versatility of the proposed method were confirmed through numerical experiments with an increased number of robots and objects, including unlearned weight objects. The effectiveness of the temporary deadlock avoidance was also confirmed by introducing additional robots within an episode. The proposed method enables the implementation of task allocation strategies that are feasible for different numbers of robots and various transport tasks without prior consideration of feasibility.
\end{abstract}

%%%%%%%%%%%%%%%%%%%%%%%%%%%%%%%%%%%%%%%%%%%%%%%%%%%%%%%%%%%%%%%%%%%%%%%%%%%%%%%%
\section{INTRODUCTION}

\begin{figure}[b]
  \centering
  \includegraphics[width=1.0\linewidth]{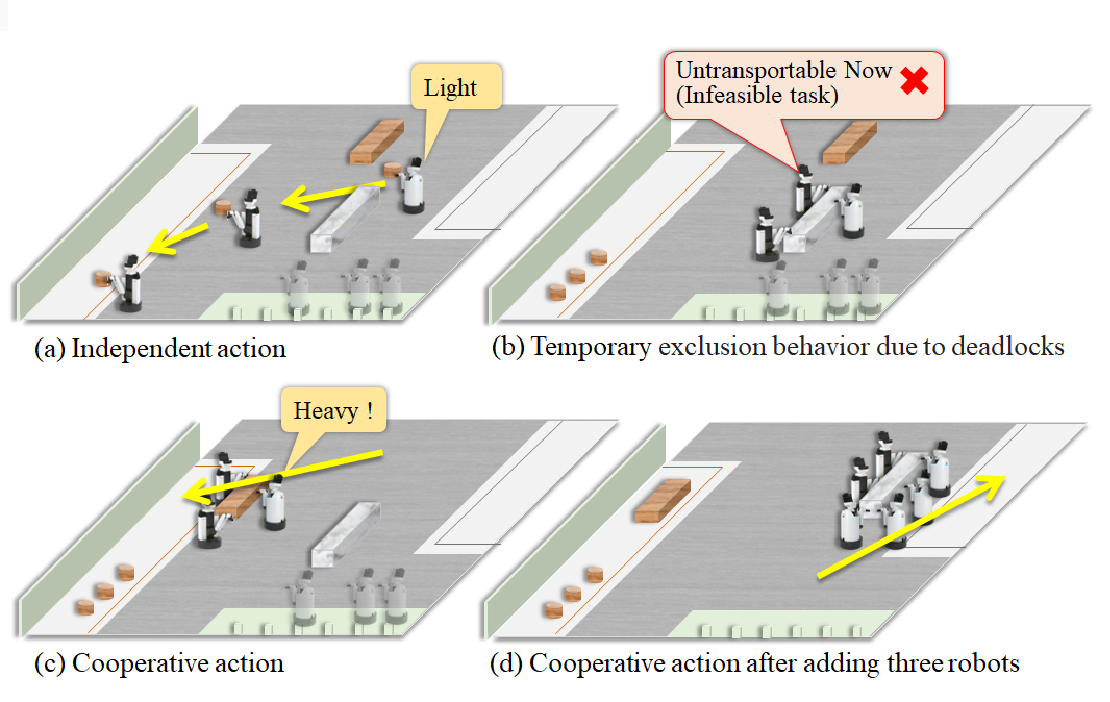}
  \caption{
    \textcolor{black}{
    Multi-object transport using a multi-robot system. (a) Robots independently perform actions when they can carry the selected objects. (b) Deadlock occurs when robots are unable to cooperatively carry the selected object. (c) Robots employ deadlock avoidance strategies and cooperate with each other to carry heavy objects. 
    \textcolor{black}{
    (d) After the introduction of three additional robots into the system, the robots cooperatively carry the object that was previously unable to be carried.
    }%textcolor
    }%textcolor
  }
  \label{fig:multi-robot cooperative}
\end{figure}

In recent years, multi-robot transportation tasks have attracted considerable attention in various fields such as delivery services, factory logistics, search and rescue, and precision agriculture. Systems in which multiple robots are controlled via a cloud server to execute various transportation tasks within facilities have been developed. These multi-robot systems achieve efficient transportation because each robot can cover a wide area independently \cite{ref:MRTA1} (Fig. \ref{fig:multi-robot cooperative} (a)). In addition, scalability is realized by enabling multiple robots to cooperate with each other when transporting objects that individual robots find infeasible \cite{ref:cooperate1,ref:cooperate2,ref:cooperate3,ref:cooperative4} (Fig. \ref{fig:multi-robot cooperative} (c)). Furthermore, distributed control improves the resilience of robots by facilitating the seamless addition of robots to the system.

Multi-robot task allocation (MRTA) is important for achieving multiple transportation tasks with multi-robot systems \cite{ref:MRTA1,ref:MRTA3}. Garkely et al. (2004) \cite{ref:MRTA1} categorized MRTA approaches into centralized algorithms \cite{ref:hungarian,ref:integer-linear-programming}, distributed algorithms \cite{ref:decentralized-path-allocation}, and hybrid algorithms that combine centralized and distributed approaches, such as auction-based methods \cite{ref:market-based,ref:decentralized-auction,ref:multi-robot-auction}.

Recently, several approaches based on multi-agent reinforcement learning (MARL) \cite{ref:MARL} have been proposed \cite{ref:Shibata}. However, MARL remains challenging because of the partial observability of each robot and the simultaneous learning of policies for each robot \cite{ref:MARL,ref:transformer}; therefore, the approach involving centralized training and decentralized execution (CTDE) is often used \cite{ref:MADDPG,ref:CTDE}. The CTDE approach has demonstrated scalability for systems with various numbers of robots that need to perform diverse tasks \cite{ref:Shibata,ref:scalable-MARL}. Furthermore, cooperative actions can be performed by learning communication between robots \cite{ref:Shibata,groups-of-robots}. 

Naturally, there are objects that cannot be transported (infeasible tasks) by all robots (Fig. \ref{fig:multi-robot cooperative} (b)), and there may be no prior information about them. In conventional methods \cite{ref:MRTA1}, the cost and task completion probability for each task must be specified explicitly. In addition, when the method \cite{ref:Shibata} based on MARL is applied to task allocation in an environment in which infeasible tasks exist, all robots are connected to the infeasible tasks, causing a stoppage (deadlock) in the robot movements.

To avoid deadlocks, it is necessary to exclude infeasible tasks from task allocation. However, when additional cooperative robots are introduced, the tasks become feasible, as shown in Fig. \ref{fig:multi-robot cooperative} (d), and it is desired to release the exclusion. Therefore, a mechanism is required to temporarily exclude the allocation of tasks %that have been decided as infeasible by past task experiences.
that have already been allocated.

In this paper, we propose a framework for dynamic task allocation that enables a multi-robot system to continue executing tasks even when there are infeasible tasks among the transportation tasks. Specifically, 
a cloud server stores the task experiences in a scalable manner with respect to the number of robots and broadcasts them to the entire robot system. In our proposed method, each robot learns the exclusion level for each object based on the task experience and other information. 
The task priorities \cite{ref:Shibata} 
are reset using an output gate based on the task experience and exclusion levels. Consequently, individual transportation, cooperative transportation, and the temporary exclusion of tasks that were considered as infeasible were achieved. 
%we build a mechanism to detect and exclude infeasible tasks, and integrate it with the mechanism \cite{ref:Shibata} that manages priorities for each transportation task. In our proposed method, each robot cooperates with neighboring robots to learn the priorities and exclusion levels of neighboring transportation tasks, enabling individual transportation. Subsequently, By learning the timing to utilize global information from a cloud server, we achieve cooperative transportation through consensus protocols that utilizes global information. Furthermore, for each transportation task, we store task experiences in a scalable manner with respect to the number of objects and robots. This allows us to detect infeasible tasks using exclusion levels and reset their priorities.
Finally, we validated the performance of our proposed method in terms of success rate and transportation time by performing numerical experiments with a larger number of robots and objects compared with the training experiments, as well as with untrained objects and a varying number of robots.

The contributions of this study are as follows:
\begin{itemize}
\setlength{\itemsep}{0cm} %
    \item We propose a task alloation framework with infeasible tasks comprising task experiences broadcasted from the cloud server and task exlusion levels learned by structured policy models. 
    %We propose a framework using MARL for task allocation with infeasible tasks, storing task experience in a scalable manner with respect to the number of robots, learning task exclusion levels for each robot through the broadcasted task experience from the cloud server, and temporary resetting task priorities.
    \item The proposed framework differs from conventional MRTA approaches that require the specification of the cost and task completion probability, in that it can temporarily exclude infeasible tasks without prior information until additional robots are introduced. 
    \item We confirm that the proposed method successfully completes feasible tasks while excluding infeasible tasks, even in numerical experiments that differ from the training experiments, including an episode where additional robots are introduced within the episode.
\end{itemize}

The remainder of this paper is structured as follows: Section II presents related work, Section III explains the problem setting, Section IV introduces the proposed method, Section V evaluates the performance of the proposed method through numerical experiments, and Section VI concludes the paper.

\section{Related Work}
\subsection{Combinatorial Optimization}
Task allocation can be formulated as a combinatorial optimization problem that aims to minimize the total travel distance, and can be solved using methods such as the Hungarian algorithm \cite{ref:hungarian} or integer linear programming \cite{ref:integer-linear-programming}. For these methods, it is necessary to provide the required resources (number of robots) for each task. However, obtaining this information in advance may not always be possible. In addition, while the tasks need to be feasible, there are cases in which the allocated resources are insufficient to execute tasks.

\subsection{Bio-inspired Approach}
Metaheuristic methods \cite{ref:insect1,ref:insect2,ref:Q-learning}, which were inspired by biological systems and natural processes, such as the division of labor in social insects, have been used to solve MRTA problems \cite{ref:MRTA3}. A commonly used approach is the threshold model \cite{ref:insect1,ref:insect2}, in which each robot selects tasks using activation thresholds and a stimulus associated with each task based on local information. These methods are flexible and can be adapted to different conditions with %different 
varying 
number of robots and tasks. However, infeasible tasks may be allocated to robots, which can lead to decreased efficiency.

\subsection{Auction Approach}
The auction algorithm is a commonly used method for task allocation among multi-robots \cite{ref:MRTA1}, and has been studied using both centralized and decentralized approaches. With the centralized approach \cite{ref:auction}, an auctioneer allocates tasks to bidders. In contrast, Choi et al. (2009) \cite{ref:decentralized-auction} proposed a decentralized auction-based algorithm without an auctioneer. With this method, tasks are allocated based on a consensus algorithm that involves local communication among bidders (robots). However, their method focused on problems in which a single robot could execute each task. 
%The auction algorithm is a commonly used method for task allocation among multi-robots \cite{ref:MRTA1}, and has been studied using both centralized and decentralized approaches. With the centralized approach \cite{ref:auction}, an auctioneer collects bids from bidders and allocates tasks to the highest bidder. In contrast, Choi et al. (2009) \cite{ref:decentralized-auction} proposed a decentralized auction-based algorithm that does not require an auctioneer. With this method, a consensus algorithm is employed to estimate the bids from other robots through local communication. The robots then allocate tasks among themselves to the highest bidder based on the estimated bids. However, their method focused on problems in which a single robot could execute each task. 

Braquet and Bakolas (2021) \cite{ref:multi-robot-auction} addressed a problem similar to that in our research. It focuses on the cases in which multiple robots are required for each task. Their method employed a consensus algorithm that is similar to that of Choi et al. (2009) \cite{ref:decentralized-auction} to estimate the lists of selected tasks, winning bids, and completed allocations. The robots allocate tasks to the robot with the highest bid among the unallocated robots based on the list of completed allocations. However, this method requires the probability of completing each task, which becomes computationally challenging when dealing with objects %with unknown weights.
for which the resources required for completion are unknown. 

\subsection{MARL Approach}
Previous studies \cite{ref:MARL-MRTA1,ref:MARL-MRTA2,ref:MARL-MRTA3,ref:MARL-MRTA4,ref:MARL-MRTA5,ref:MARL-MRTA6} have focused on task allocation problems using MARL. These methods formulate task allocation as a Markov decision process and learn policies using learning algorithms, such as MADDPG \cite{ref:MADDPG}. However, these methods %learn with
are designed for 
a fixed number of agents and tasks, rendering them ineffective under different conditions. To address this issue, policy models that only obtain neighboring agents and tasks have been utilized \cite{ref:Shibata,ref:scalable-MARL}. In particular, a framework for task allocation in the presence of an unknown number of robots required for cooperative transportation was proposed in previous studies \cite{ref:Shibata}. This proposed framework utilizes dynamic priorities for each task and global robot communication. However, if the tasks do not have sufficient resources (number of robots), the robots may encounter a deadlock scenario until the introduction of additional robots.

Our proposed framework, which is similar to the approach proposed by Shibata et al. (2022) \cite{ref:Shibata}, uses dynamic priorities, but in contrast to their method, it incorporates the learning of dynamic exclusion levels for each task by leveraging task experiences broadcasted from the cloud server. Dynamic exclusion levels are used in the output gate, which temporarily resets dynamic priorities. Therefore, if a robot encounters a deadlock with a specific object, the priority of that task is reset. In addition, when more cooperative robots are introduced, the reset for that object is released, enabling the robots to again be transported cooperatively.

\section{Problem Setting}
Consider the task of transporting $M$ different objects with varying weights using $N$ robots to the goals associated with the tasks (Fig. \ref{fig:multi-robot cooperative}). However, the weights of the objects are unknown and there may be objects that cannot be transported. In this study, the transport task aims to efficiently transport all feasible objects using all robots as quickly as possible.

The study focuses on the allocation of objects to robots. Therefore, the robot allocated to object $l$ reaches it within the shortest possible time. The number of robots allocated to object $l$, denoted by $|\mathcal{C}_l|$, determines the ability of robots to carry objects. If $|\mathcal{C}_l|$ is greater than the weight $w_l$ of object $l$, a sufficient number of robots are connected to the object. Subsequently, the object is transported to the goal $\bm{z}^*_l\in\mathbb{R}^2$. Here, 
\begin{equation*}%\label{eq:number of robot}
 \mathcal{C}_l=\{i\in\{1,\cdots,N\}|\|\bm{x}_i-\bm{z}_l\|\le\delta\}
\end{equation*}
represents the set of robots that have actually connected to object $l$, where $\bm{x}_i\in\mathbb{R}^2$ is the position of robot $i(=1,\cdots,N)$, $\bm{z}_l\in\mathbb{R}^2$ is the position of object $l$. If distance between robots and objects is smaller than small positive constant $\delta$, the robots are able to connect to the object.

This study made the following assumptions: 
\begin{itemize}
\setlength{\itemsep}{0cm} 
    \item The cloud server can always obtain the latest information (global information) about the robots and objects. 
    \item \textcolor{black}{
    When cooperating with other robots is necessary, each robot can obtain global information via the cloud server.
    }%textcolor
\end{itemize}

\section{Method}
\begin{figure}[t]
  \centering
  \includegraphics[width=1.0\linewidth]{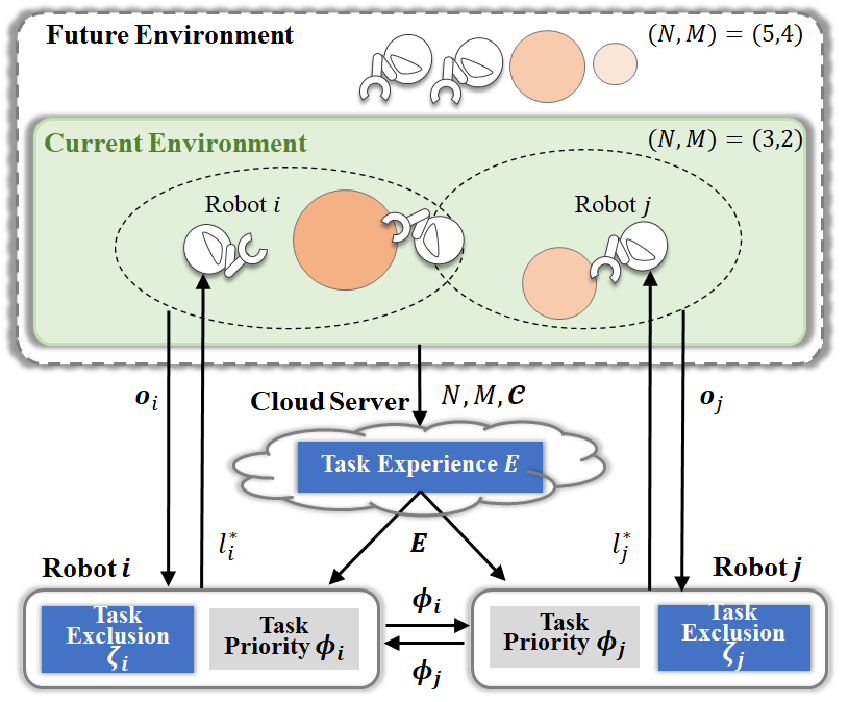}
  \caption{
   Abstracted framework of the proposed dynamic task allocation, which includes handling infeasible tasks. 
   \textcolor{black}{
   Task experiences $E$ that are scalable with the number of robots $N$, are broadcasted from the cloud server to each robot in order to learn task exclusion levels $\zeta_i$ and task priorities $\phi_i$. Subsequently, robots shares their information between all robots when cooperating is necessary. 
   }%textcolor
   }
  \label{fig:framework}
\end{figure}

The weights of the objects are unknown, and there may be objects that cannot be transported; hence, the simple allocation of tasks to the robots in advance would result in a deadlock. 
Therefore, we propose the framework shown in Fig. \ref{fig:framework}. %consisting of %four 
%three components for each robot:
%, based on the extension of the dynamic priorities by K.~Shibata et al. (2022) \cite{ref:Shibata}: 
%\begin{itemize}
%\setlength{\itemsep}{0cm} %
%    \item Task Experience $E$ %Policy Network 
%    (Section \ref{sec:NN})
%    \item Dynamic Task Exclusion
%    $\zeta_i$
%    (Section \ref{sec:DynamicExclusion})
%    \item Integration with Dynamic Task Priority \cite{ref:Shibata}
%    $\phi_i$
%    (Section \ref{sec:DynamicPriority})
    %\item Output Gate (Section\ref{sec:OutputGate})
%\end{itemize}
The cloud server holds global information, such as the task experience $E$ for the entire robot system. In addition, each robot $i\in\{1,\cdots,N\}$ has a task exclusion level $\zeta_i^l$ for each object $l\in\{1,\cdots,M\}$ in addition to the task priority $\phi_i^l$. %Robot $i$ learns the timing to utilize the information from the cloud server and the exclusion levels and priorities of the neighboring $K(<M)$ objects $l\in\mathcal{N}^\mathrm{Load}_i$ using the static policy network model. 
The task exclusion level enables the temporary resetting of the task priority for each object based on the broadcasted $E$ from the cloud server. Consequently, deadlock avoidance was enabled.
%For infeasible objects, the output gate, which take the exclusion level in account, is used to reset the priorities.

\subsection{Task Experience}\label{sec:NN}
%In this paper, to achieve scalable policies for the number of robots and objects, each robot $i$ utilizes partial observables, denoted as follows:
The cloud server stores the overall task experience $E_l(t)$ for object $l$ in a scalable manner based on the number of robots. $E_l(t)$ can be expressed as follows:
\begin{equation}\label{eq:E}
  E_l(t)=\int_0^t 
  \left( \frac{\bigl(1-\sigma_0(|\bm{v}_l(\tau)|)\bigr)\cdot|\mathcal{C}_l(\tau)|}{N}\right) ^\kappa\,d\tau.
\end{equation}
Here, $|\mathcal{C}_l(t)|$ represents the number of robots connected to object $l$ at current time $t$, and $\bm{v}_l\in\mathbb{R}^2$ represents the velocity of object $l$. In addition, $\sigma_0$ is a step function with a threshold of $0$, and $\kappa$ is a positive constant. 

\subsection{Dynamic Task Exclusion}\label{sec:DynamicExclusion}
To obtain exclusion levels that are scalable with the number of objects $M$, robot $i$ learns a static policy network model that outputs the target exclusion level ${\zeta_i^l}^\ast$ for $K(<M)$ neighboring objects $l\in\mathcal{N}^\mathrm{Load}_i$ as follows:
\begin{equation}\label{eq:zeta_i_l}
{\zeta_i^l}^\ast = \pi_i^\zeta(o_i).
\end{equation}
Robot $i$ sets the current exclusion levels $\zeta_i^l$ for other objects. The target exclusion levels for all objects $l(=\{1,\cdots,M\})$ are set as follows:
\begin{equation*}
  {d_i^l}=
    \begin{cases}
    {\displaystyle 
    {\zeta_i^l}^\ast
    },
    \hspace{3.5mm} 
    \text{ $l\in\mathcal{N}^\mathrm{Load}_i$} \\
    {\displaystyle 
     \zeta_i^l
    },
    \hspace{6.5mm} 
    \text{otherwise}
  \end{cases}.
\end{equation*}
Each robot $i$ utilizes partial observables $\bm{o}_i$, denoted as follows: 
\begin{eqnarray}\label{oi}
\bm{o}_i \!\!\!&=&\!\!\! [ \bm{x}_i^\top, \bm{\phi}_i^{l_{i1}}, \cdots, \bm{\phi}_i^{l_{iK}}, \nonumber\ \\
& & 
\bm{x}_{j_{i1}}^\top, \bm{\phi}_{j_{i1}}^{l_{i1}}, \cdots, \bm{\phi}_{j_{i1}}^{l_{iK}}, 
\cdots, 
\bm{x}_{j_{iK}}^\top, \bm{\phi}_{j_{iK}}^{l_{i1}}, \cdots, \bm{\phi}_{j_{iK}}^{l_{iK}}, \nonumber\ \\
& & 
\bm{z}_{l_{i1}}^\top, {\bm{z}_{l_{i1}}^\ast}^\top, \bm{v}_{l_{i1}}^\top, 
\cdots, 
\bm{z}_{l_{iK}}^\top, {\bm{z}_{l_{iK}}^\ast}^\top, \bm{v}_{l_{iK}}^\top, \nonumber\ \\
& &
{E}_{l_{i1}}, \cdots, {E}_{l_{iK}} ]^\top,
\end{eqnarray}
which contains $K(<M)$ nearest robots $j\in\mathcal{N}^\mathrm{Robot}_i:=\{j_{i1},\cdots,j_{iK}\}$, objects $l\in\mathcal{N}^\mathrm{Load}_i:=\{l_{i1},\cdots,l_{iK}\}$, and task experience $E_l$ for the nearest objects.
%During training, parameters of the network model $\pi_i^\zeta$ and $\pi_i^\beta$ are centralized updated. During execution, each robot performs decentralized execution using only its own observations and the learned network parameters (CTDE).

%However, in Equation (\ref{phi1}), it is not possible to quickly change the exclusion levels for objects outside the neighborhood. It is desired to efficiently release resources (number of robots) from objects with task experiences that could not be executed. Therefore, we add the following consensus protocol to Equation (\ref{eq:zeta_i_l}), sharing the exclusion levels of other robots at the necessary timing $\sigma(\beta_i)$ via the cloud server: 

Furthermore, for objects $l\notin\mathcal{N}^\mathrm{Load}_i$ that are outside the $K(<M)$ nearest neighbors of robot $i$, a mechanism is introduced to share the exclusion levels with other robots via the cloud server at time when $\sigma(\beta_i)$. The timing was obtained by learning a static policy network model, which is denoted as follows: 
\begin{equation}\label{eq:beta_i}
\beta_i = \pi_i^\beta(o_i)\in[0,1].
\end{equation}

%Then, 
Based on the above, robot $i$ updates the exclusion levels for all objects $l(=\{1,\cdots,M\})$ using a consensus protocol via the cloud server:
%\begin{equation}\label{phi1}
% \dot{\zeta_i^l} = k_\zeta({d_i^l}-\zeta_i^l)
%\end{equation}
\begin{equation}\label{eq:zeta_hat}
 \dot{\zeta_i^l} = k_\zeta({d_i^l}-\zeta_i^l)
 +k_\zeta\sigma(\beta_i) N (\max_{j} \zeta_j^l-\zeta_i^l).
\end{equation}
Here, $k_\zeta$ is a positive constant,
\begin{equation*}
  \sigma(x)=
    \begin{cases}
    1, & \text{$x\ge 0.5$} \\
    0, & \text{otherwise}
    \end{cases}
\end{equation*}
is a step function with a threshold of $0.5$ for variable $x\in[0,1]$.

\subsection{Integration with Dynamic Task Priority}\label{sec:DynamicPriority}
Each robot sequentially allocates objects using dynamic priorities \cite{ref:Shibata}.

Similar to the exclusion levels, to achieve scalable learning with a number of objects $M$, robot $i$ sets the target priorities of all objects $l(=\{1,\cdots,M\})$
as follows: 
\begin{equation*}
  c_i^l=
    \begin{cases}
    {\displaystyle 
    {\phi_i^l}^\ast
    },
    \hspace{3.5mm} 
    \text{ $l\in\mathcal{N}^\mathrm{Load}_i$} \\
    {\displaystyle 
     \phi_i^l
    },
    \hspace{6.5mm} 
    \text{otherwise}
  \end{cases},
\end{equation*}
%${\phi_i^l}^\ast$  for the $K(<M)$ neighboring objects $l\in\mathcal{N}^\mathrm{Load}_i$ using a policy network model, and sets the current value for other objects 
where $\phi_i^l$ is the current priority of robot $i$ for object $l$; in addition, robot $i$ learns the target priorities ${\phi_i^l}^\ast$ for $K(<M)$ neighboring objects $l\in\mathcal{N}^\mathrm{Load}_i$ using a policy network model as follows: 
\begin{equation}\label{eq:phi_i_l}
{\phi_i^l}^\ast=\pi_i^\phi(o_i).
\end{equation}

Cooperation with robots that are not located nearby is required for handling heavy objects. Therefore, similar to the exclusion level, when the required timing is 
$\sigma(\alpha_i)$, robot $i$ updates the priorities for all objects $l(=\{1,\cdots,M\})$ using a consensus protocol via a cloud server: 
\begin{equation}\label{eq:phi_hat}
 \dot{\phi_i^l} = k_\phi({c_i^l}-\phi_i^l)
 +k_\phi\sigma(\alpha_i)\sum_{j=1}^N (\phi_j^l-\phi_i^l).
\end{equation}
Here, $k_\phi$ denotes a positive constant. 
In addition, the timing for sharing priorities with other robots is obtained by learning a static policy network as follows: 
\begin{equation}\label{eq:alpha_i}
\alpha_i = \pi_i^\alpha(o_i)\in[0,1].
\end{equation}

\begin{figure}[t]
  \centering
  \includegraphics[width=1.0\linewidth]{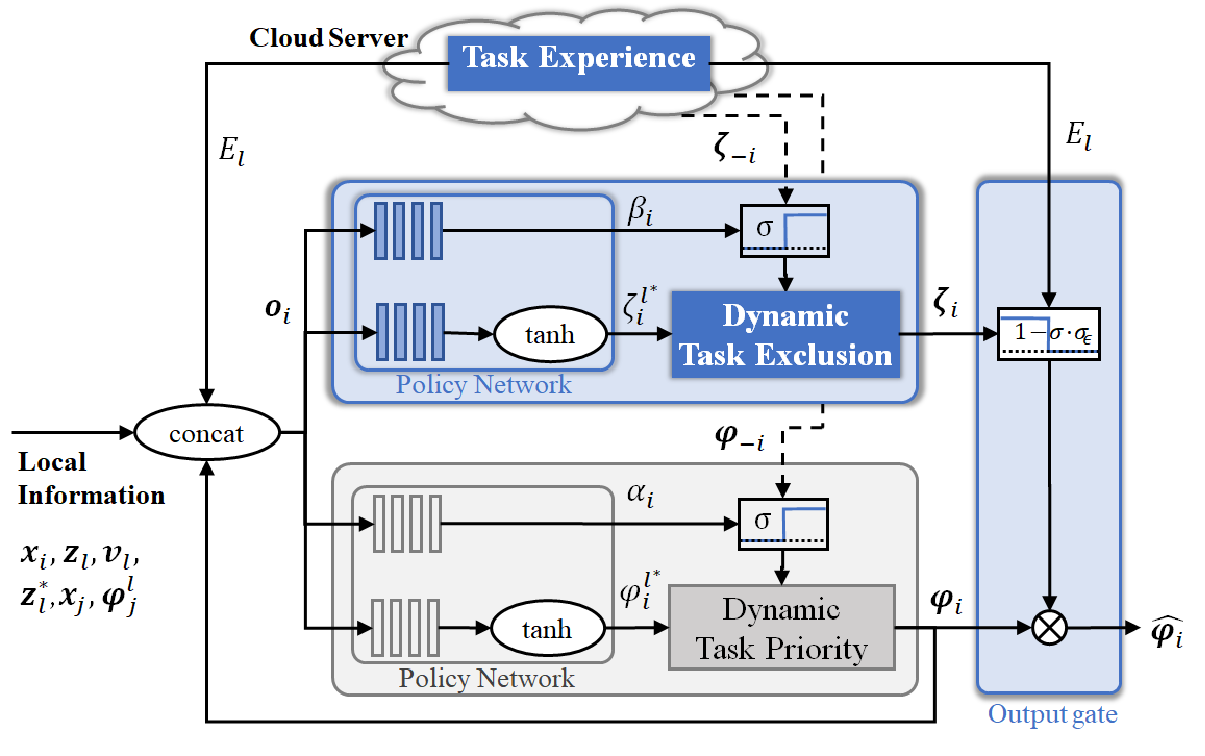}
  \caption{
   Block diagram of multi-robot task allocation with infeasible tasks. \textcolor{black}{The output gate plays a role in integrating dynamic task exclusion and dynamic task priority.} 
   }
  \label{fig:diagram}
\end{figure}

The exclusion level $\zeta_i^l$ in Equation (\ref{eq:zeta_hat}) is integrated with priority $\phi_i^l$ in Equation (\ref{eq:phi_hat}) using the output gate, as shown in Fig. \ref{fig:diagram}. The integration is expressed as follows:
\begin{equation}\label{eq:hat_phi}
  \hat{\phi_i^l}=
  \left( 1-\sigma(\zeta_i^l)\cdot\sigma_\mathrm{\epsilon}(E_l)\right) \phi_i^l.
\end{equation}
Here, $\sigma_\mathrm{\epsilon}$ is a step function with the threshold of a small positive constant $\mathrm{\epsilon}$. If object $l$ satisfies $\zeta_i^l \ge 0.5$ and $E_l \ge \mathrm{\epsilon}$, transportation is determined to be infeasible with the current resources, and its priority is reset.

Finally, robot $i$ selects the object with the highest priority among priorities $\hat\phi_i^l\ 
(l=1,\cdots,M)$ after the output gate in Equation (\ref{eq:hat_phi}):
\begin{equation*}
 l_i^\ast=\underset{l\in\{1,2,\cdots,M\}} {\operatorname{arg max}} \hat{\phi}_i^l.
\end{equation*} 
Furthermore, the priorities of the objects that have reached their goals or are being transported by other robots are set to $0$. In addition, if an object is already allocated and is being transported, the update of all priorities is stopped.

\subsection{Policy Optimization}\label{sec:Policy}
As shown in Fig. \ref{fig:diagram}, using Equation (\ref{eq:zeta_i_l}), (\ref{eq:beta_i}), (\ref{eq:phi_i_l}), and (\ref{eq:alpha_i}), a static policy network model of robot $i$ is as follows: 
\begin{equation}\label{eq:policy}
\bm{a}_i = 
\mathrm{\pi}_i(\bm{o}_i).
\end{equation}
Here, $\bm{a}_i$ is
\begin{equation*}
\bm{a}_i
=\begin{bmatrix}
{\phi_i^{l_{i1}}}^\ast,\cdots,{\phi_i^{l_{iK}}}^\ast, \alpha_i,  {\zeta_i^{l_{i1}}}^\ast,\cdots,{\zeta_i^{l_{iK}}}^\ast, \beta_i 
\end{bmatrix}^\top.
\end{equation*}
In this study, the MADDPG algorithm \cite{ref:MADDPG} was used to perform the centralized training of the policy network model of Equation (\ref{eq:policy}).

%To learn the $N$ static policy network models possessed by $N$ robots, the MADDPG algorithm developed by R.~Lowe et al. (2017) \cite{ref:MADDPG} is used. MADDPG uses a centralized training and decentralized execution (CTDE) approach that considers decentralized execution. During training, the network model parameters were updated by maximizing the discounted cumulative reward over a single episode with observations of all the robots and the reward received when the objects reached their respective goals, as defined in the reward function $r$ \cite{ref:Shibata}. During execution, each robot performs decentralized execution using only its own observations and learned network parameters.

\section{Numerical Experiment}
In this section, we present the results of numerical experiments on the transportation tasks of multiple objects. The performance of the proposed method was evaluated under various settings by varying the number of robots and objects.

\subsection{Simulation Setup}
\begin{figure}[t]
  \centering
  \includegraphics[width=0.8\linewidth]{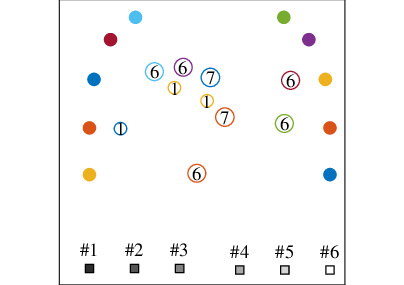}
  \caption{
  \textcolor{black}{
  Numerical environments with $N=6$ robots and $M=10$ objects. Each square represents a robot, each colored circle represents an object, and each colored dot represents a goal of transport. The number inside each circle represents the number of robots needed to transport the object, and the number of each square represents the robot's ID.
  }%textcolor
  }
  \label{fig:environment}
\end{figure}

%\subsection{Experimental Conditions}\label{sec:Experiment}
Starting positions of %the robots and 
the participants were randomly selected, 
and the target positions of each object were fixed for each simulation (Fig. \ref{fig:environment}). The carrying capacity of each robot was set to $1$. By the inclusion of $N$ robots cooperating to transport the same object, it becomes possible to transport an object with a weight of $N$.

In the training experiments (Training in Table \ref{Table:Number}), $N(=3)$ robots and $M(=6)$ objects were used. The weights of the objects were set to three types, including objects that cannot be transported by $N$ robots, with weights $w_l\in\{1,N,N+1\}$. Among the six objects, three objects had a weight of $N + 1$, whereas the remaining three objects were randomly chosen from weights of $1$ and $N$ with a $50\ \%$ probability.

\begin{table}[b]
    \centering
    \caption{
    \textcolor{black}{
    Number of objects and robots in the experiments 
    %The numbers in parentheses indicate the probability of appearance. In experiment of Validation 3, the number of objects and robots change to the number indicated by the arrows in the table.
    }%textcolor
    }
    \label{Table:Number}
    \begin{tabular}{cccccccc}
        \hline
        \multirow{2}{*}{Experiment} & 
        Robot & 
        Object &
        \multicolumn{5}{c}{Object weight} \\ 
        \cline{4-8}
        & $N$ & $M$ & $1$ & $3$ & $4$ & $6$ & $7$ \\ 
        \hline \hline
        Training & $3$ & $6$ & $[0\ 3]$ & $[0\ 3]$ & $3$ & $0$ & $0$ \\
        Validation 1 & $6$ & $10$ & $[0\ 8]$ & $[0\ 8]$ & $2$ & $0$ & $0$ \\
        Validation 2  & $6$ & $10$ & $[0\ 8]$ & $0$ & $0$ & $[0\ 8]$ & $2$ \\
        Validation 3  & $3\rightarrow6$ & $10$ & $1$ & $4$ & $0$ & $5$ & $0$ \\
        \hline
    \end{tabular}
\end{table}

%\subsection{Training Conditions}
We used the MADDPG code \cite{ref:github} and set the simulation parameters listed in Table \ref{Table:parameters}. %The reward received when the objects reach to their goals, as defined in reward function $r$ \cite{ref:Shibata}. 
In the proposed method, the numbers of neighboring robots and objects for each robot were set to $K=2$. The other parameters were set as follows: $\kappa=10$, $\mathrm{\epsilon}=1$, $k_\phi=0.2$, $k_\zeta=0.2$.

%When the number of robots $N$ increases, the probability of Policy Gradient updating the policy in the correctly direction decreases exponentially \cite{ref:MADDPG}. In such cases, it is necessary to increase the number of calculations and carefully perform updates. Although the number of robots was not large, the dimensionality of the robot actions in the proposed method was high; therefore, the learning rate was set to 1/10 of the value in the code \cite{ref:github} ($0.001$). The batch size of the learning data per update was set to $1024$ episodes, as in \cite{ref:github}.

\begin{table}[t]
    \centering
    \caption{Simulation parameters.}
    \label{Table:parameters}
    \begin{tabular}{cc}
        \hline
        Parameter & Value \\ \hline \hline
        Sampling period [s] & 1.0\\
        Number of steps per episode & 300\\
        Number of episodes & 50,000\\
        Number of hidden layers (critic) & 4\\
        Number of hidden layers (actor) & 4\\
        Activation function (hidden) & RelU\\
        Activation function (output, critic) & tanh\\
        Activation function (output, actor) & linear\\
        Optimizer (critic) & Adam\\
        Optimizer (actor) & SGD\\
        Discount factor & 0.99\\
        Batch size & 1024\\
        \hline
    \end{tabular}
\end{table}

The validation experiments (refer to validations 1 and 2 in Table \ref{Table:Number}) used $N(=6)$ robots and $M(=10)$ objects. Among these, two objects have an untransportable weight of $w_l=N+1(=7)$, whereas the remaining eight objects have weights of either $\{1,3\}$ or $\{1,6\}$. The proportion of objects with weights of $3$ or $6$ was selected using $\{0, 50, 100\ \%\}$ probabilities, and $100$ episodes were run, with each episode consisting of 1,000 steps. 

To evaluate the scalability and %generalized performance 
versatility of the proposed method, we used the following two criteria:
\begin{itemize}
\setlength{\itemsep}{0cm} %
    \item Success rate: The percentage of episodes in which all objects reached their goals within the total number of steps.
    \item Transportation time: The average number of steps (seconds) until all transportable objects reached their goals for the episodes in which they succeed in transportation.
\end{itemize}

Furthermore, in an additional validation experiment (refer to Validation 3 in Table \ref{Table:Number}), the number of robots was varied from $3$ to $6$ during each episode to evaluate the effectiveness of the proposed method. 

\subsection{Training Performance}%{Centralized Training}
\begin{figure}[t]
  \centering
  \includegraphics[width=1.0\linewidth]{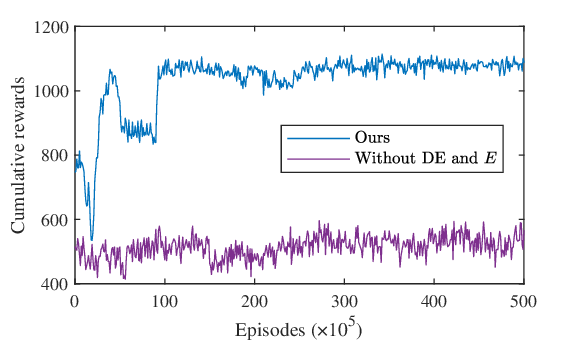}
  \caption{
  Cumulative rewards of our framework using DP: Dynamic Priority and DE: Dynamic Exclusion, and $E$: task experience of objects.
  }
  \label{fig:rewards}
\end{figure}

The training performance of the proposed method was evaluated based on the average cumulative reward over $3$ training runs. The results were compared with those of a method that used only the dynamic priority mechanism \cite{ref:Shibata} without DE and $E_l$), as shown in Fig. \ref{fig:rewards}. Here, DE in the figure refers to dynamic exclusion.

The average cumulative reward of the method using only the dynamic priority mechanism \cite{ref:Shibata} was low because the robots became stuck in a deadlock while gathering around objects that could not be transported.

However, the proposed method achieved high cumulative rewards and did not cause deadlocks. It is believed that the output gate in Equation (\ref{eq:hat_phi}), which is controlled by the dynamic exclusion in Equation (\ref{eq:zeta_hat}), effectively avoids deadlock.

\subsection{Scalability and Versatility}
\begin{figure}[t]
  \centering
  \includegraphics[width=1.0\linewidth]{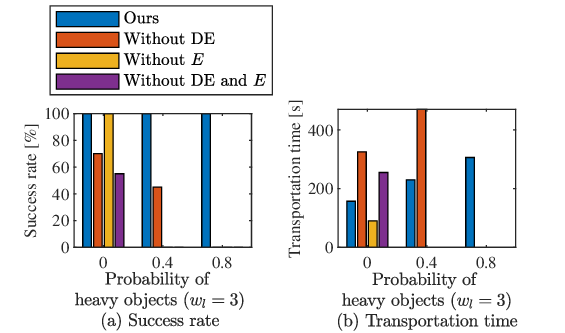}
  \caption{Validation results of each learned distributed policy with 6 robots and 10 objects with weights of $w_l\in\{1,3,7\}$.}
  \label{fig:validation}
%\end{figure}

%\begin{figure}[h]
  \centering
  \includegraphics[width=1.0\linewidth]{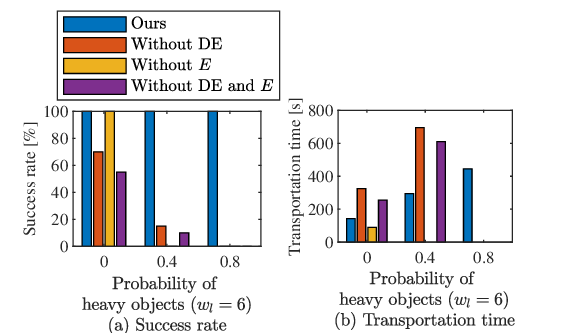}
  \caption{Validation results of each learned distributed policy with 6 robots and 10 objects with weights of $w_l\in\{1,6,7\}$.}
  \label{fig:validation2}
\end{figure}

%{\textcolor{black}{Scalability and Generalized Performance Validation}}%{Decentralized Execution}
Fig. \ref{fig:validation} shows the validation results obtained when increasing the number of robots and objects compared to the learning phase, with the weights of the objects remaining the same. For comparison, the results excluding the dynamic exclusion (DE) and $E$ inputs from the proposed method are shown in red and yellow, respectively. When using only the dynamic priority mechanism \cite{ref:Shibata} without DE and $E$, there were cases in which objects could not be transported because of deadlock occurrences with infeasible objects, resulting in a success rate below $100\ \%$. However, the proposed method achieved successful transportation in shorter times than the other methods under all conditions. Furthermore, the proposed method reduced the transportation time when there were fewer heavy objects. This indicates that both individual and cooperative transportation occurred simultaneously, depending on the composition of the objects. These results demonstrated the scalability of the proposed method.

Fig. \ref{fig:validation2} shows the validation results obtained when increasing the number of robots and objects compared to the learning phase, including the case where there are unlearned weight values for the objects. Even for objects with unlearned weights ($=6$), the proposed method successfully completed the transport tasks. This demonstrates the versatility of the proposed method.

In Fig. \ref{fig:validation} and \ref{fig:validation2}, the policies obtained from the learning of methods removing either the DE or $E$ input from the proposed method resulted in a significant decrease in the success rate. For the method without DE (shown by the red bars in the figures), even when the ratio of weights $3$ to $6$ is zero (i.e., all objects except for infeasible ones can be transported individually with weight $1$), deadlock occurs because of the infeasible objects, and the success rate does not reach 100\ \%. For the method without $E$ input (shown by the yellow bars in the figures), when the ratio of weights $3$ to $6$ was $0$, the success rate reached $100\ \%$. However, when heavy objects are transported cooperatively, the success rate decreases to $0\ \%$. From these results, it can be inferred that the removal of DE results in the loss of deadlock avoidance functionality, and the removal of the $E$ input resulted in the loss of cooperative transportation capability.

\subsection{Effectiveness of Temporary Priority Exclusions}
%{\textcolor{black}{Validating the Temporal Effectiveness of the DE}}

\begin{figure}[t]
  \centering
  \includegraphics[width=1.0\linewidth]{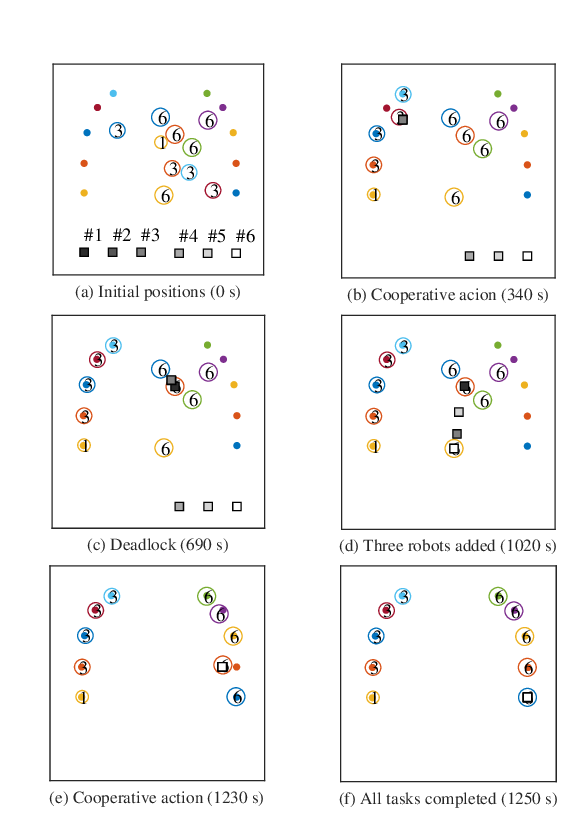}
  \caption{
  \textcolor{black}{
  Validation results of proposed method with $10$ objects and after varying the number of robots from $3$ to $6$.
  }%textcolor
  }
  \label{fig:add.experiment}
\end{figure}

\begin{figure}[t]
  \centering
  \includegraphics[width=1.0\linewidth]{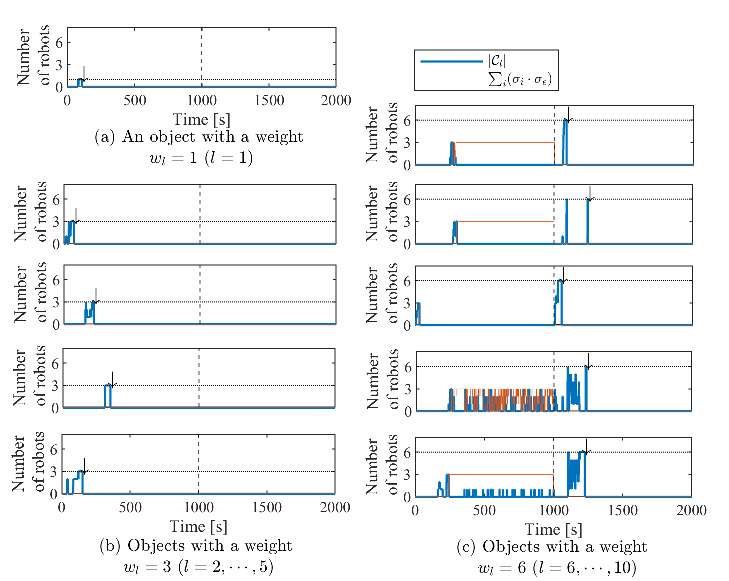}
  \caption{
  \textcolor{black}{
  Time series data of number of robots allocated to objects $|\mathcal{C}_l|$ and task exclusion signals $\sigma_i\cdot\sigma_\epsilon$. Dotted and dashed lines show the number of robots required to transport each object and the timing when three robots are added. The arrows pointing down indicate the time when each object is transported to each goal.
  }%textcolor
  }
  \label{fig:add.history}
\end{figure}

\begin{figure}[t]
  \centering
  \includegraphics[width=1.0\linewidth]{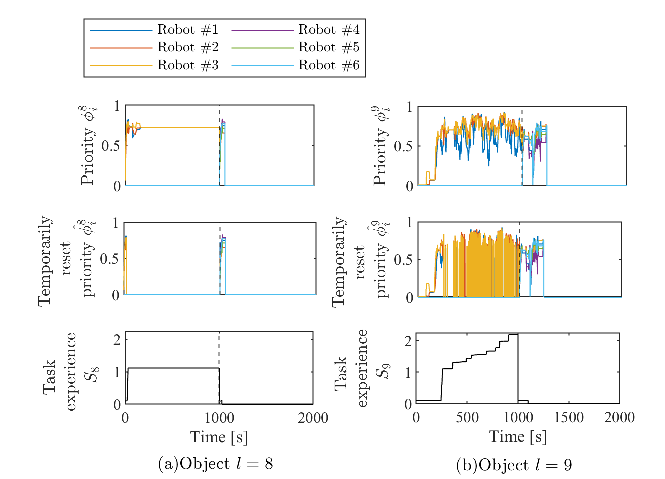}
  \caption{
  \textcolor{black}{
  Time series data of priorities $\phi_i^l$, $\hat{\phi_i^l}$ and task experience $E_l$, ($i=1,\cdots,6$, $l=8,9$).
  %Time history of priorities $\phi_i^l$, $\hat{\phi_i^l}$ and task experience $E_l$ of tasks with IDs $l=1,3$.
  }%textcolor
  }
  \label{fig:add.Sl}  
\end{figure}

We confirmed the effect of temporarily resetting the priority using output gates through an experiment (Validation 3) in which robots were added within an episode. Fig. \ref{fig:add.experiment} shows the results sampled during a $2,000$ s episode. In Fig. \ref{fig:add.experiment} (a)--(c), $3$ cooperating robots were able to transport objects weighing up to $3$ before encountering a deadlock with objects weighing $6$. In Fig. \ref{fig:add.experiment} (d)--(f), with $3$ additional robots, all robots cooperate to transport all objects weighing $6$.

Fig. \ref{fig:add.history} shows the number of robots connected to objects $l=1,\cdots,10$ as $|\mathcal{C}_l|$, and the values of the output gates $\sigma_i\cdot\sigma_\mathrm{\epsilon}$ for each robot $i=1,\cdots,6$. The results of the output gates are shown in Fig. \ref{fig:add.history} (c); evidently, the priority of objects weighing $6$ was temporarily excluded. Consequently, they were able to transport objects weighing $1$ and $3$ to the goal early without encountering a deadlock. Furthermore, after the addition of $3$ robots, the output gates were released, allowing the $6$ robots to collaboratively transport objects weighing $6$.

Fig. \ref{fig:add.Sl} shows the task experiences $E_8$ and $E_9$ for objects $l=8$ and $l=9$ with a weight of $6$. After $3$ robots were added at $1,000$ s, the current task experiences were recalculated with $N=6$ and the values decreased. Consequently, the temporary exclusions were released. Normalization and recalculation of task experiences using current number of robots $N$ enables the proposed method to handle objects with a weight of $6$.

\section{CONCLUSIONS}
In this study, we propose a framework for dynamic task allocation in multiple transportation tasks where the weights of the objects are unknown and there are infeasible objects. In the proposed method, multi-robots sequentially select objects using a dynamic task allocation approach. To achieve scalability in terms of the number of robots and objects, we learn the timing for utilizing global information and the target priorities of neighboring tasks based on local observations. We calculated dynamic priorities using an consensus protocol. We also constructed target exclusion levels and consensus protocols to temporarily reset priorities using an output gate, thereby avoiding deadlocks caused by infeasible objects. This construction enables task allocation strategies to be executable without prior consideration of the feasibility of the task. 
By performing numerical experiments, we confirmed the efficiency of task allocation through both individual and cooperative transportation, which can be achieved even with an increasing number of robots and objects (scalability), the ability to allocate tasks to unlearned objects (versatility), and 
the effectiveness of the temporary 
prevention of deadlocks 
while the number of robots was insufficient to execute the tasks. 
%even in the presence of infeasible objects. 

Our proposed method assumes the use of a cloud server to obtain the required global information. In the future, we plan to implement the proposed method to validate its  feasibility in real-world environments.

\addtolength{\textheight}{-13cm}   % This command serves to balance the column lengths


%%%%%%%%%%%%%%%%%%%%%%%%%%%%%%%%%%%%%%%%%%%%%%%%%%%%%%%%%%%%%%%%%%%%%%%%%%%%%%%%

%%%%%%%%%%%%%%%%%%%%%%%%%%%%%%%%%%%%%%%%%%%%%%%%%%%%%%%%%%%%%%%%%%%%%%%%%%%%%%%%

%%%%%%%%%%%%%%%%%%%%%%%%%%%%%%%%%%%%%%%%%%%%%%%%%%%%%%%%%%%%%%%%%%%%%%%%%%%%%%%%
\begin{comment}
\section*{APPENDIX}

Appendixes should appear before the acknowledgment.

\section*{ACKNOWLEDGMENT}

The preferred spelling of the word ÒacknowledgmentÓ in America is without an ÒeÓ after the ÒgÓ. Avoid the stilted expression, ÒOne of us (R. B. G.) thanks . . .Ó  Instead, try ÒR. B. G. thanksÓ. Put sponsor acknowledgments in the unnumbered footnote on the first page.

%%%%%%%%%%%%%%%%%%%%%%%%%%%%%%%%%%%%%%%%%%%%%%%%%%%%%%%%%%%%%%%%%%%%%%%%%%%%%%%%

References are important to the reader; therefore, each citation must be complete and correct. If at all possible, references should be commonly available publications.
\end{comment}

\bibliography{arXiv}

\bibliographystyle{IEEEtran}

\end{document}